\documentclass[sigconf]{acmart}

\AtBeginDocument{%
  \providecommand\BibTeX{{%
    \normalfont B\kern-0.5em{\scshape i\kern-0.25em b}\kern-0.8em\TeX}}}

\usepackage{subcaption}
\usepackage{amsmath}
\makeatletter
\newcommand*\bigcdot{\mathpalette\bigcdot@{.7}}
\newcommand*\bigcdot@[2]{\mathbin{\vcenter{\hbox{\scalebox{#2}{$\m@th#1\bullet$}}}}}
\makeatother

\copyrightyear{2021} \acmYear{2021} \setcopyright{rightsretained} \acmConference[CODS COMAD 2021]{8th ACM IKDD CODS and 26thCOMAD}{January 2--4, 2021}{Bangalore, India}
\acmBooktitle{8th ACM IKDD CODS and 26th COMAD (CODS COMAD2021), January 2--4, 2021, Bangalore,India}\acmDOI{10.1145/3430984.3430986}\acmISBN{978-1-4503-8817-7/21/01}

\begin{document}

\title{ActiveNet: A computer-vision based approach to determine lethargy}
\author{Aitik Gupta}
\affiliation{\institution{ABV-IIITM, Gwalior}}
\email{aitikgupta@gmail.com}

\author{Aadit Agarwal}
\affiliation{\institution{ABV-IIITM, Gwalior}}
\email{agarwal.aadit99@gmail.com}

\begin{abstract}

The outbreak of COVID-19 has forced everyone to stay indoors, fabricating a significant drop in physical activeness. Our work is constructed upon the idea to formulate a backbone mechanism, to detect levels of activeness in real-time, using a single monocular image of a target person. The scope can be generalized under many applications, be it in an interview, online classes, security surveillance, et cetera.

We propose a Computer Vision based multi-stage approach, wherein the pose of a person is first detected, encoded with a novel approach, and then assessed by a classical machine learning algorithm to determine the level of activeness. An alerting system is wrapped around the approach to provide a solution to inhibit lethargy by sending notification alerts to individuals involved.
\end{abstract}


\maketitle

\section{Introduction}
Activeness, both physical and mental, has been one of the primary healthcare concerns since the inception of smart devices. It has undergone a significant drop, especially since the COVID-19 outbreak \cite{mckibbin2020global}. This is due to lazy and sedentary practices, either during leisure time or while working from home.

Despite recent advancements in Computer Vision and related domains, reliable security surveillance systems with little to no human intervention \cite{4906450} is still a challenge, especially now that these times call for unfortunate motivations and impulses.

There has been significant research on drowsiness detection using Computer Vision \cite{6144162}, but most of them leverage the degree to which the person's eyes are open or closed. While this setting is instrumental for in-car webcams, they fail for long-range distances, which is a standard paradigm in CCTV footages, security webcams, et cetera.

Moreover, numerous papers related to the study of body language using Computer Vision pertain to emotion detection \cite{article}; wherein there has been an apparent \textit{lack of attention, given to attention.}

With the outlook of a more generalized solution, we demonstrate a multi-stage mechanism to identify activeness of a person, such as in online classes, job interviews, security surveillance systems, et cetera, aiming to rectify diminished activeness in students, interviewees, security guards respectively. The input to the pipeline would just be a single RGB image. To realize lethargy in this type of multi-stage approach, an intermediate representation of information is essential. At the end of the first stage, we maintain a 2-dimensional Cartesian plane coordinate information of various joints of a person. Just using this representation, it is almost impractical to achieve our goal, which drives the idea of our novel pose encoding stage, where we maintain an angular representation of data, aiming to remove all positional aspects in the data. With abstraction after every stage, the resulting system becomes more robust to the inherent \textit{noise} of the data, such as visual characteristics of various people, different coordinates for different image views or sizes, et cetera.

\section{Architecture}
ActiveNet is a machine-learning based system, the proposed architecture for which accepts a camera input. It is then operated through a Human Pose Estimation module to extract keypoints of the joints in human body. The pose encoding module generates a vector, which is calculated by the angles of certain joints. The angles are imputed and scaled via intermediaries generated while training a self-scraped dataset, discussed later. After processing, a traditional machine-learning classifier takes the angles as input. The classifier predicts the activeness level, based on which the alert system is triggered.

\begin{figure*}[ht]
\centering
\includegraphics[width=0.7\textwidth, height=0.35\textwidth]{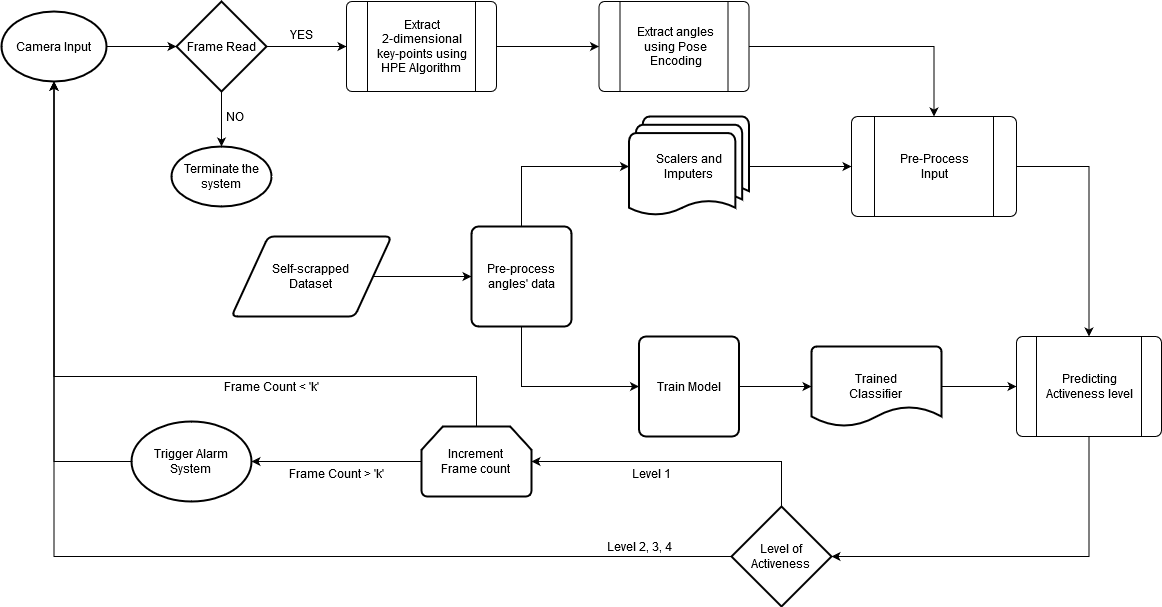}
\vspace{-0.25cm}
\caption{ActiveNet Architecture}
\label{Fig1}
\end{figure*}

\subsection{Human Pose Estimation}\label{hpe}
\subsubsection{Literature Review}
Human Pose Estimation (HPE) is one of the key aspects in computer vision that has undergone tremendous research in the last few years. Its numerous applications are one of the main reasons for the importance it has gained. It estimates the configuration of the body (pose) from a single, typically monocular, image. To study the methods of estimating the position of joints from an image, one must understand some of the most significant challenges: 
\begin{itemize}
    \item Background noise, lighting, visibility issues
    \item High variance of visual appearance and physique of humans
    \item Partial occlusions due to self-articulation and layering of objects
    \item Complexity of human skeletal structure and its hierarchy
    \item Information loss by projecting a 3D object to a 2D plane
\end{itemize}
Some classical approaches are based on a pictorial-structure framework. Early works introduced a mixture model of parts, which expresses joint relationships \cite{6380498}. This approach has the limitation of having the pose model independent of image data. Later on, deep learning based regression approaches were introduced \cite{A}, which brought a shift in the research paradigm towards them. Most of the later researches operate over convolutional building blocks, and have been universally adopted for image-data driven approaches. Gradually, direct keypoint regression-based methods were replaced by more promising heatmap regression methods \cite{7780880}. 

Broadly classified, there are two approaches to the convolutional architectures for Single Person Pose Estimation (SPPE) or Multi-Person Pose Estimation (MPPE). The first approach, the \textbf{top-down} approach, is decoupled into two sub-problems. Firstly, a person detection algorithm is claimed, followed by a pose estimation algorithm for every detected person. State-of-the-art (SOTA) solutions for the sub-problems could potentially be utilized together in the pipeline. The inference speed of this approach strongly depends on the number of detected people inside an image. The second approach, called the \textbf{bottom-up} approach, is more resilient to the number of people in an image, and thereby, could potentially be faster than the first approach. Firstly, all possible keypoints are detected in an image, followed by grouping by human instances.

\subsubsection{Our Work}
We based our work on the popular bottom-up method Lightweight OpenPose \cite{osokin2018lightweight_openpose}, for mainly two reasons. Firstly, this work heavily optimizes the original OpenPose \cite{8765346} implementation to reach real-time inference speeds on CPU with negligible accuracy drop, which can further be optimized for Edge Devices using Intel\textregistered{} OpenVINO\texttrademark{} Toolkit, as done in other researches \cite{intel_openvino}. Secondly, in the context of finding the levels of alertness in a person, our work is dependent on the angles generated by different body joints. Therefore, the \textbf{bottom-up} approach is well suited for isolation of joints, even in cases where other parts are not detected. We propose a pre-processing step for the pose encoding, which takes care of the joints for which HPE module could not determine the 2D locations of keypoints. It will be discussed later in the paper.

We leverage the pre-trained weights provided by open-sourced library at: \href{https://github.com/Daniil-Osokin/lightweight-human-pose-estimation.pytorch}{https://github.com/Daniil-Osokin/lightweight-human-pose-estimation.pytorch}, which contains the implementation of LightWeight OpenPose \cite{osokin2018lightweight_openpose}. The weights are trained over the COCO dataset \cite{10.1007/978-3-319-10602-1_48}, consisting of 17 keypoints in the following order: (1) Nose; (2) Neck; (3) Right Shoulder; (4) Right Elbow; (5) Right Wrist; (6) Left Shoulder; (7) Left Elbow; (8) Left Wrist; (9) Right Hip; (10) Right Knee; (11) Right Ankle; (12) Left Hip; (13) Left Knee; (14) Left Ankle; (15) Right Eye; (16) Left Eye; (17) Right Ear; (18) Left Ear. The Neck (2) keypoint is just a 2-dimensional mean of Right Shoulder (3) and Left Shoulder (6) keypoints in the Cartesian coordinate system, therefore making it a total of 17 keypoints, along with one additional extrapolated keypoint.

\subsection{Pose Encoding}\label{pe}
\begin{figure}[ht]
 \begin{subfigure}{.2\textwidth}
  \includegraphics[width=1\linewidth]{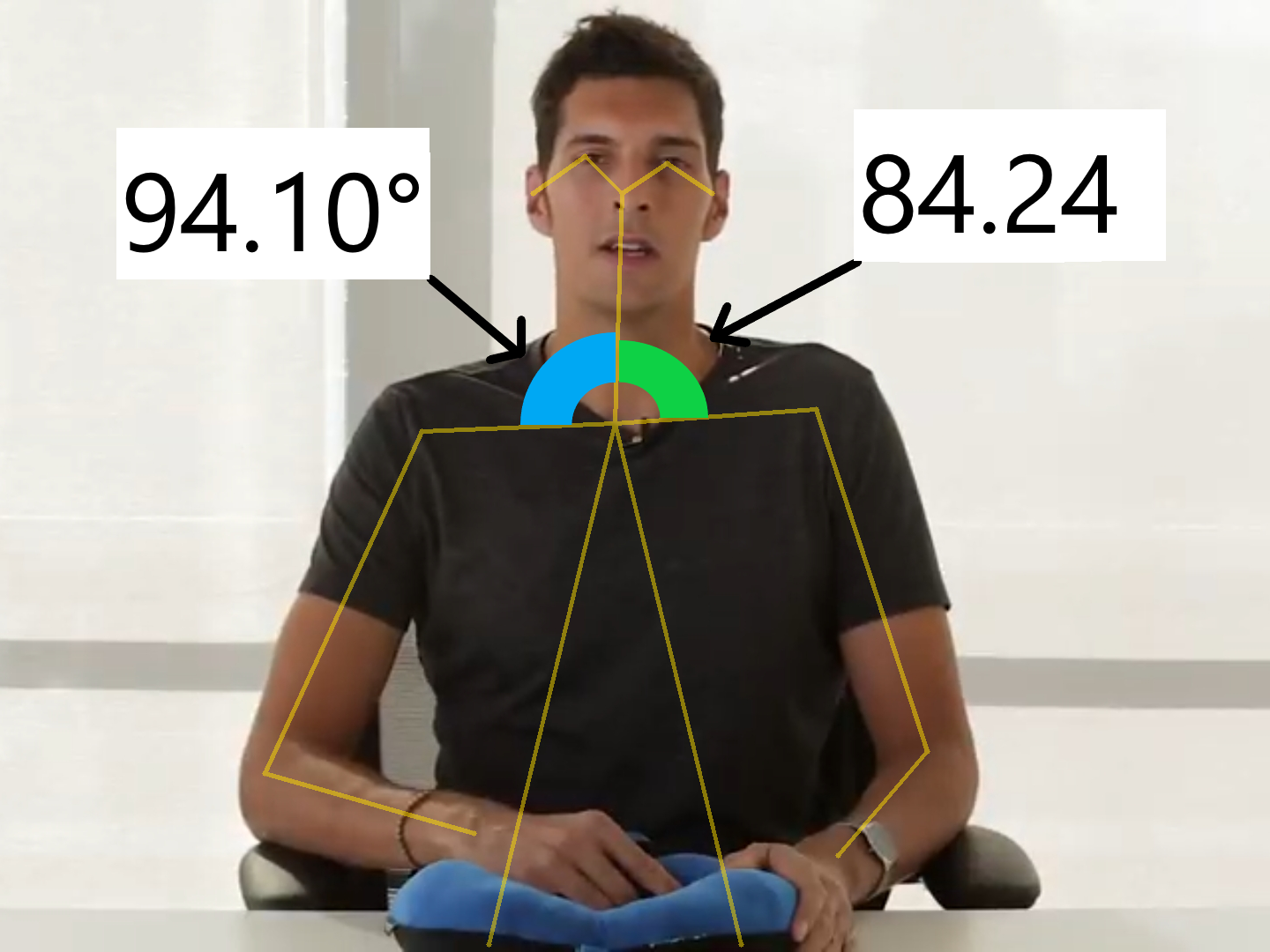}
  \caption{High activeness}
  \label{straight}
 \end{subfigure}
 \begin{subfigure}{.2\textwidth}
  \includegraphics[width=1\linewidth]{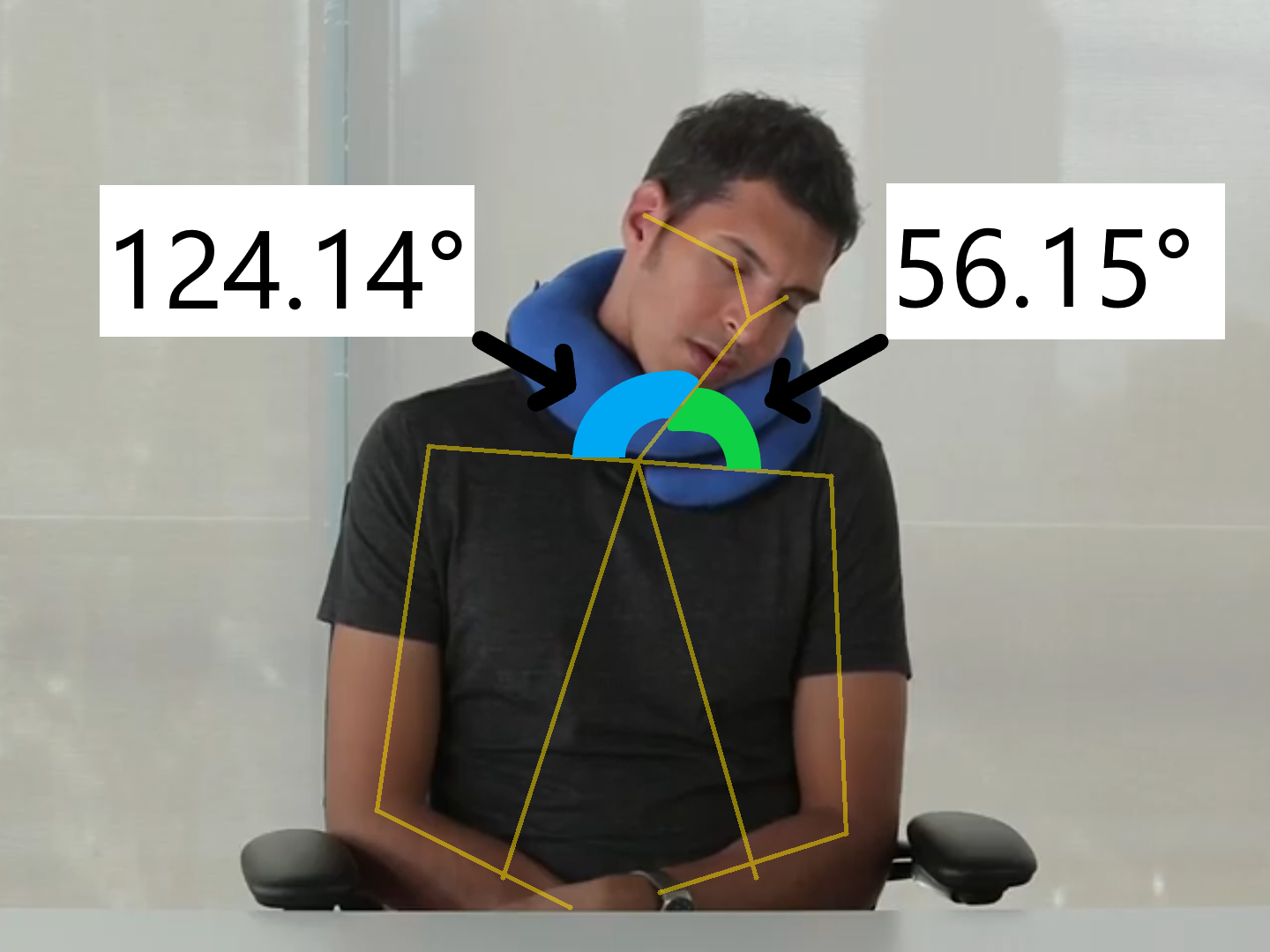}
  \caption{Low activeness}
  \label{bent}
 \end{subfigure}
\vspace{-0.25cm}
\caption{Angular Encoding}
\label{encoding}
\end{figure}

We propose a pose encoding technique, with the aim of removing any positional configurations from the detected keypoints. We claim this by taking the angles between different subsets of joints in the upper and lower body. Regardless of the position of a person in an image, the absolute values of angles between the joints remain same for a single type of pose. We do not use the keypoints directly, as \autoref{position} explains that even if the images are just a mirror of each other and the contextual pose of the person remains same, their estimations can be quite different. Those are dependent on the positional aspects in the camera input. A slight change in the camera's offset would lead to an entirely different set of 2D keypoint estimations. Another motivation for angular representation can be explained by \autoref{encoding}; as the angle between shoulders, neck and head changes, the abstract activeness of the person changes along with it. We consider the angles made by following subset of joints:
\begin{figure}[ht]
 \begin{subfigure}{0.4\linewidth}
  \centering
  \includegraphics[width=0.75\linewidth]{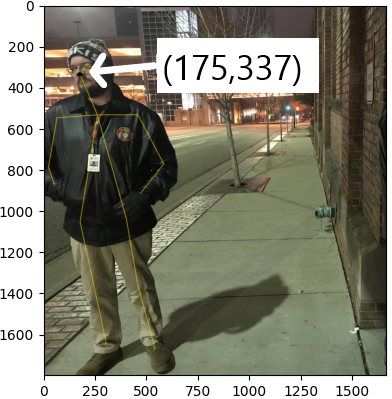}
  \caption{Original Image}
  \label{left}
 \end{subfigure}
 \begin{subfigure}{0.4\linewidth}
  \centering
  \includegraphics[width=0.75\linewidth]{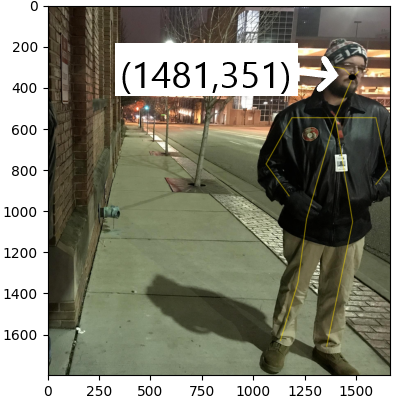}
  \caption{Mirrored Image}
  \label{right}
 \end{subfigure}
\vspace{-0.25cm}
\caption{Nose keypoint estimation with augmentation}
\label{position}
\end{figure}

\begin{enumerate}
    \item Nose, Eyes, Ears
    \item Neck, Nose, Ears
    \item Shoulders, Neck, Nose
    \item Neck, Shoulders, Elbows
    \item Shoulders, Elbows, Wrists
    \item Nose, Neck, Core Joint
    \item Neck, Core Joint, Hips
    \item Hips, Knees, Ankles
\end{enumerate}
The subset is chosen such that it captures the possible range of motions in most sections of human body, and thus, numerous poses. We consider Core Joint as the 2D mean of Left Hip and Right Hip joints in the Cartesian plane, and the actual angles are calculated using dot product.

Encoding a single pose creates an array of 15 elements, containing angular data in degrees. The joints which are not detected in the pose, are represented by (-1,-1) in the coordinate axes. We give \textbf{NaN} values in the encoding when two out of three keypoints (from \textbf{A}, \textbf{B} and \textbf{C}) have the same 2D coordinates, which includes the (-1,-1) case. This can well occur due to occlusions in the image, or semi-accurate estimations. If this is not handled during encoding, a \textbf{ZeroDivisionError} error is raised. \textbf{NaN} values are later handled in \autoref{pp}.
Since this module inputs just 2D coordinates of joints detected by HPE algorithm discussed in \autoref{hpe}, it is invariant to the visual aspects of the image, inherently expanding the generalizability to new image-data.
\vspace{-0.45cm}

\subsection{Dataset}\label{data}
To train our model on the encoding from the module discussed in \autoref{pe}, we scraped images from web using different keywords for different classes. The dataset contains 4 classes for 4 levels of activeness-zones:
\begin{enumerate}
    \item Level 1: Below 25\%
    \item Level 2: Between 25-50\%
    \item Level 3: Between 50-75\%
    \item Level 4: Above 75\%
\end{enumerate}
There are 40 images for each class, scraped using keywords such as "army soldiers" for Level 4 activeness, while "sleeping while standing" for level 1 activeness.

\begin{figure*}[ht]
 \begin{subfigure}{.21\textwidth}
  \includegraphics[width=.8\linewidth]{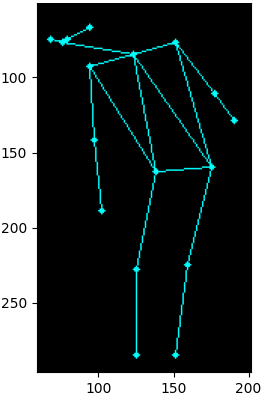}
  \caption{Level 1}
  \label{75}
 \end{subfigure}
 \begin{subfigure}{.21\textwidth}
  \includegraphics[width=.8\linewidth]{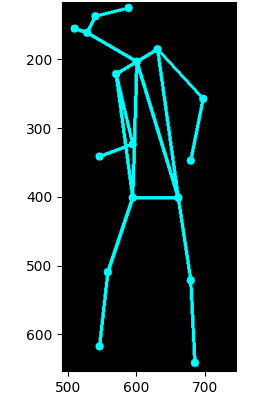}
  \caption{Level 2}
  \label{50_75}
 \end{subfigure}
 \begin{subfigure}{.21\textwidth}
  \includegraphics[width=.8\linewidth]{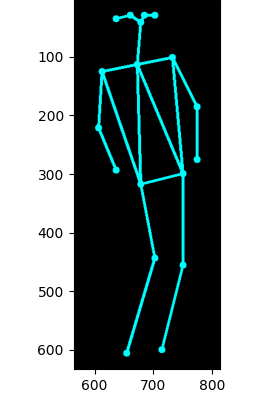}
  \caption{Level 3}
  \label{25_50}
 \end{subfigure}
 \begin{subfigure}{.21\textwidth}
  \includegraphics[width=.8\linewidth]{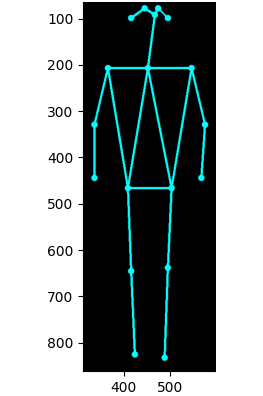}
  \caption{Level 4}
  \label{25}
 \end{subfigure}
\vspace{-0.25cm}
\caption{Activeness Levels}
\label{levels}
\end{figure*}

\subsection{Pre-Processing}\label{pp}
\subsubsection{Treating \textbf{NaN} Values}
Pose encoding yields \textbf{NaN} values as discussed in \autoref{pe}. Treating these values is a prominent discussion in data science. One approach is to eliminate the encoding which has more than half \textbf{NaN} values. Another approach is to use numpy.ma module \cite{group} which provides a convenient way to address this issue, by introducing masked arrays, which are either no-mask representing only clean entries or boolean arrays indicating presence of invalid entries, in which case the invalid entries are eliminated, allowing the classifier to train on only valid entries. The approach of pose encoding as described in \autoref{pe} and as shown in \autoref{encoding} makes this possible, even with fewer features.

\subsubsection{Scaling Features}
The raw features from the encodings are scaled down to a similar range across the whole dataset. This helps in many ways during training, as priority can be given to optimizing the weights based on the correlation of features to the target rather than scales of various features. We use the standard scaler provided by scikit-learn for this module.

\subsection{Training}\label{ml}
To train a classifier using the scaled data discussed in \autoref{pp}, we adapt the ensemble algorithms based on Decision Trees \cite{10.1023/A:1016409317640}, such as Random Forest Classifier (RFC) \cite{10.1023/A:1010933404324} and XGBoost Classifier \cite{Chen:2016:XST:2939672.2939785}. We first train and validate a Logistic Regression algorithm \cite{ref1} to establish the benchmark scoring for classification. Subsequently we tune the hyper-parameters using scikit-learn's GridSearchCV. The best scoring results are then chosen for each algorithm, while using K-Fold cross-validation technique. \autoref{clf-results} provides algorithm-wise results. 
\subsection{Evaluation and Analysis}\label{eval}

\begin{table}[h]
\caption{Classifier-Wise Results}\label{clf-results}
  \begin{tabular}{|l|c|}
    \toprule
    \textbf{Classifier}&\textbf{Accuracy}\\
    \midrule
    Logistic Regression&56.67\%\\
    Decision Tree Classifier&66.67\%\\
    XGBoost Classifier&63.34\%\\
    Random Forest Classifier (RFC)&76.67\%\\
  \bottomrule
\end{tabular}
\end{table}
\begin{table}[h]
\caption{Class-Wise Results for RFC}\label{class-result}
  \begin{tabular}{|l|c|c|c|}
    \toprule
    \textbf{Class}&\textbf{Precision}&\textbf{Recall}&\textbf{F1-Score}\\
    \midrule
    {Level 1}&0.989&0.820&0.896\\
    {Level 2}&1.000&0.745&0.853\\
    {Level 3}&0.914&0.663&0.768\\
    {Level 4}&0.989&0.807&0.888\\
    \bottomrule
\end{tabular}
\end{table}

Based on experimentation, the best results are achieved using RFC with an accuracy of 76.67\%. As \autoref{encoding} explains, even a small subset of joints can be adequate to assess the overall activeness. Hence a general interpretation can be lined that tree-based ensemble algorithms would do better here, as the individual trees are built on subsets of features.  A detailed evaluation of the classifier results is given in \autoref{class-result}. Analysing the results, we find that the extreme classes have the best scores compared to the intermediate classes.  \textbf{Our self-scraped dataset does not represent an ideal distribution of the real-world classes, the outcomes so achieved only provide a baseline solution for our approach.}

\subsection{Alert Mechanism}\label{alert_mech}
At the time of inference, an alerting mechanism provides a method to inform people in-concern if diminished levels of activeness is developed in the targeted person. This is done in faith to keep the targeted person agile. To get rid of false positives, and in order to make the pipeline more robust, we raise a notification only when \textit{k} number of contiguous frames are classified in the lowest class, where \textit{k} is an arbitrary constant. A reasonable value of \textit{k} can change with the domain. The alert module relies on Slack Workspaces. Other works have used the same utility, but for different use cases \cite{doi:10.1002/aps3.11280} \cite{thesis_outlier}.

We configure a \textit{notification-alert bot} after enabling the \textbf{Incoming Webhook} functionality in Slack Workspace:
\begin{enumerate}
    \item Webhook is a unique URL to send HTTP requests
    \item Only organizations with the Webhook URL can send alerts
    \item Only organizations in the workspace can receive alerts
    \item The alert can be sent with a customized message, along with a time-stamp
    \item Works on all platforms, requires Slack to be installed
\end{enumerate}
For demonstration purposes, we create a demo Slack Workspace \textbf{active-networkspace}, typically meant for an organization. Here for example, it is meant for \textit{active-net} organization.
\begin{figure}[h]
\includegraphics[width=0.7\linewidth, height=0.5\linewidth]{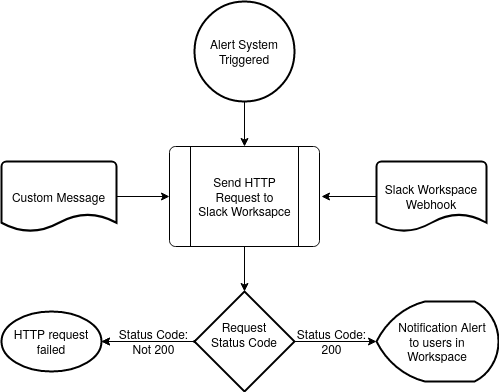}
\vspace{-0.25cm}
\caption{Alert Mechanism Flowchart}
\label{Fig2}
\end{figure}
\section{Demonstration}
For demonstration, we use a desktop webcam, rigged with a single Nvidia GeForce GTX 1650 graphics card to run the whole pipeline, inferencing at around 34 frames per second. The code, along with screenshots are available at: \href{https://github.com/aaditagarwal/ActiveNet}{github.com/aaditagarwal/ActiveNet}

\section{Current Limitations}
As there exists no well defined dataset for such type of classification task, in addition to discussion in \autoref{eval}, we end up with weights which do not generalize for \textit{unseen poses}. One of the main complication arises when camera view angle shifts. For instance, a situation where the person is looking sideways, the keypoint estimations will be from a totally different distribution compared to estimations when person is looking directly at the camera. Consecutively, angles between joints do not provide a good encoding solution in this situation. Considering the limitations of the self-scraped dataset, we overlooked the implications of such issues.

Another limitation to be realized is the loss of information projecting a 3-dimensional object to a 2-dimensional space. The 2D keypoint estimations, and their corresponding encoding, both suffer from this limitation. Considering the recent developments in 3D pose estimation, which are either end-to-end \cite{sun2020multi}, i.e. operate on a monocular image, or take 2D estimations from HPE modules and lift them to 3 dimensional space \cite{zhaoCVPR19semantic}, theoretically, encodings could be improved further to support angles made in 3 dimensions. Given our aim for a real-time solution, we decided not to address the 3 dimensional approach, but can be approached in future.

\section{Conclusion}
Through this paper, we explained our multi-stage approach to identify levels of activeness with a novel pose encoding stage. Once trained on enough poses, it can potentially be used in any domain without retraining, given that the HPE module is robust enough to generate the keypoints. We demonstrated the idea with a working pipeline, using Slack to alert users in a workspace with custom messages. We welcome future research in this domain with a strong belief that \textbf{Activeness should be given active attention}, both mental and physical.

\bibliographystyle{ACM-Reference-Format}
\bibliography{sample-base}


\begin{thebibliography}{20}

%%% the Web address.

\ifx \showCODEN    \undefined \def \showCODEN     #1{\unskip}     \fi
\ifx \showDOI      \undefined \def \showDOI       #1{#1}\fi
\ifx \showISBNx    \undefined \def \showISBNx     #1{\unskip}     \fi
\ifx \showISBNxiii \undefined \def \showISBNxiii  #1{\unskip}     \fi
\ifx \showISSN     \undefined \def \showISSN      #1{\unskip}     \fi
\ifx \showLCCN     \undefined \def \showLCCN      #1{\unskip}     \fi
\ifx \shownote     \undefined \def \shownote      #1{#1}          \fi
\ifx \showarticletitle \undefined \def \showarticletitle #1{#1}   \fi
\ifx \showURL      \undefined \def \showURL       {\relax}        \fi
% invisible to TeX
\providecommand\bibfield[2]{#2}
\providecommand\bibinfo[2]{#2}
\providecommand\natexlab[1]{#1}
\providecommand\showeprint[2][]{arXiv:#2}

\bibitem[\protect\citeauthoryear{{Assari} and {Rahmati}}{{Assari} and
  {Rahmati}}{2011}]%
        {6144162}
\bibfield{author}{\bibinfo{person}{M.~A. {Assari}} {and} \bibinfo{person}{M.
  {Rahmati}}.} \bibinfo{year}{2011}\natexlab{}.
\newblock \showarticletitle{Driver drowsiness detection using face expression
  recognition}. In \bibinfo{booktitle}{\emph{2011 IEEE International Conference
  on Signal and Image Processing Applications (ICSIPA)}}.
  \bibinfo{pages}{337--341}.
\newblock


\bibitem[\protect\citeauthoryear{Breiman}{Breiman}{2001}]%
        {10.1023/A:1010933404324}
\bibfield{author}{\bibinfo{person}{Leo Breiman}.}
  \bibinfo{year}{2001}\natexlab{}.
\newblock \showarticletitle{Random Forests}.
\newblock \bibinfo{journal}{\emph{Mach. Learn.}} \bibinfo{volume}{45},
  \bibinfo{number}{1} (\bibinfo{date}{Oct.} \bibinfo{year}{2001}),
  \bibinfo{pages}{5–32}.
\newblock
\showISSN{0885-6125}
\urldef\tempurl%
\url{https://doi.org/10.1023/A:1010933404324}
\showDOI{\tempurl}


\bibitem[\protect\citeauthoryear{{Cao}, {Hidalgo Martinez}, {Simon}, {Wei}, and
  {Sheikh}}{{Cao} et~al\mbox{.}}{2019}]%
        {8765346}
\bibfield{author}{\bibinfo{person}{Z. {Cao}}, \bibinfo{person}{G. {Hidalgo
  Martinez}}, \bibinfo{person}{T. {Simon}}, \bibinfo{person}{S. {Wei}}, {and}
  \bibinfo{person}{Y.~A. {Sheikh}}.} \bibinfo{year}{2019}\natexlab{}.
\newblock \showarticletitle{OpenPose: Realtime Multi-Person 2D Pose Estimation
  using Part Affinity Fields}.
\newblock \bibinfo{journal}{\emph{IEEE Transactions on Pattern Analysis and
  Machine Intelligence}} (\bibinfo{year}{2019}), \bibinfo{pages}{1--1}.
\newblock


\bibitem[\protect\citeauthoryear{Chen and Guestrin}{Chen and Guestrin}{2016}]%
        {Chen:2016:XST:2939672.2939785}
\bibfield{author}{\bibinfo{person}{Tianqi Chen} {and} \bibinfo{person}{Carlos
  Guestrin}.} \bibinfo{year}{2016}\natexlab{}.
\newblock \showarticletitle{{XGBoost}: A Scalable Tree Boosting System}. In
  \bibinfo{booktitle}{\emph{Proceedings of the 22nd ACM SIGKDD International
  Conference on Knowledge Discovery and Data Mining}} (San Francisco,
  California, USA) \emph{(\bibinfo{series}{KDD '16})}.
  \bibinfo{publisher}{ACM}, \bibinfo{address}{New York, NY, USA},
  \bibinfo{pages}{785--794}.
\newblock
\showISBNx{978-1-4503-4232-2}
\urldef\tempurl%
\url{https://doi.org/10.1145/2939672.2939785}
\showDOI{\tempurl}


\bibitem[\protect\citeauthoryear{Connolly}{Connolly}{2018}]%
        {thesis_outlier}
\bibfield{author}{\bibinfo{person}{Aidan Connolly}.}
  \bibinfo{year}{2018}\natexlab{}.
\newblock \showarticletitle{Automated Outlier Detection in Crime Data Using
  Programming}.
\newblock \bibinfo{journal}{\emph{Undergraduate Honors Thesis, University of
  Nebraska-Lincoln}} (\bibinfo{year}{2018}).
\newblock


\bibitem[\protect\citeauthoryear{Grindstaff, Mabry, Blischak, Quinn, and
  Chris~Pires}{Grindstaff et~al\mbox{.}}{2019}]%
        {doi:10.1002/aps3.11280}
\bibfield{author}{\bibinfo{person}{Brandin Grindstaff},
  \bibinfo{person}{Makenzie~E. Mabry}, \bibinfo{person}{Paul~D. Blischak},
  \bibinfo{person}{Micheal Quinn}, {and} \bibinfo{person}{J. Chris~Pires}.}
  \bibinfo{year}{2019}\natexlab{}.
\newblock \showarticletitle{Affordable remote monitoring of plant growth in
  facilities using Raspberry Pi computers}.
\newblock \bibinfo{journal}{\emph{Applications in Plant Sciences}}
  \bibinfo{volume}{7}, \bibinfo{number}{8} (\bibinfo{year}{2019}),
  \bibinfo{pages}{e11280}.
\newblock
\urldef\tempurl%
\url{https://doi.org/10.1002/aps3.11280}
\showDOI{\tempurl}
\showeprint{https://bsapubs.onlinelibrary.wiley.com/doi/pdf/10.1002/aps3.11280}


\bibitem[\protect\citeauthoryear{Group}{Group}{[n.d.]}]%
        {group}
\bibfield{author}{\bibinfo{person}{UHC Group}.}
  \bibinfo{year}{[n.d.]}\natexlab{}.
\newblock \bibinfo{title}{Missing data: masked arrays¶}.
\newblock
\newblock
\urldef\tempurl%
\url{https://currents.soest.hawaii.edu/ocn_data_analysis/_static/masked_arrays.html}
\showURL{%
\tempurl}


\bibitem[\protect\citeauthoryear{{Ko}}{{Ko}}{2008}]%
        {4906450}
\bibfield{author}{\bibinfo{person}{T. {Ko}}.} \bibinfo{year}{2008}\natexlab{}.
\newblock \showarticletitle{A survey on behavior analysis in video surveillance
  for homeland security applications}. In \bibinfo{booktitle}{\emph{2008 37th
  IEEE Applied Imagery Pattern Recognition Workshop}}. \bibinfo{pages}{1--8}.
\newblock


\bibitem[\protect\citeauthoryear{Kristiani, Yang, and Huang}{Kristiani
  et~al\mbox{.}}{2020}]%
        {intel_openvino}
\bibfield{author}{\bibinfo{person}{Endah Kristiani}, \bibinfo{person}{Chao-Tung
  Yang}, {and} \bibinfo{person}{Chin-Yin Huang}.}
  \bibinfo{year}{2020}\natexlab{}.
\newblock \showarticletitle{iSEC: An Optimized Deep Learning Model for Image
  Classification on Edge Computing}.
\newblock \bibinfo{journal}{\emph{IEEE Access}}  \bibinfo{volume}{PP}
  (\bibinfo{date}{02} \bibinfo{year}{2020}), \bibinfo{pages}{1--1}.
\newblock
\urldef\tempurl%
\url{https://doi.org/10.1109/ACCESS.2020.2971566}
\showDOI{\tempurl}


\bibitem[\protect\citeauthoryear{Lin, Maire, Belongie, Hays, Perona, Ramanan,
  Doll{\'a}r, and Zitnick}{Lin et~al\mbox{.}}{2014}]%
        {10.1007/978-3-319-10602-1_48}
\bibfield{author}{\bibinfo{person}{Tsung-Yi Lin}, \bibinfo{person}{Michael
  Maire}, \bibinfo{person}{Serge Belongie}, \bibinfo{person}{James Hays},
  \bibinfo{person}{Pietro Perona}, \bibinfo{person}{Deva Ramanan},
  \bibinfo{person}{Piotr Doll{\'a}r}, {and} \bibinfo{person}{C.~Lawrence
  Zitnick}.} \bibinfo{year}{2014}\natexlab{}.
\newblock \showarticletitle{Microsoft COCO: Common Objects in Context}. In
  \bibinfo{booktitle}{\emph{Computer Vision -- ECCV 2014}},
  \bibfield{editor}{\bibinfo{person}{David Fleet}, \bibinfo{person}{Tomas
  Pajdla}, \bibinfo{person}{Bernt Schiele}, {and} \bibinfo{person}{Tinne
  Tuytelaars}} (Eds.). \bibinfo{publisher}{Springer International Publishing},
  \bibinfo{address}{Cham}, \bibinfo{pages}{740--755}.
\newblock
\showISBNx{978-3-319-10602-1}


\bibitem[\protect\citeauthoryear{McKibbin and Fernando}{McKibbin and
  Fernando}{2020}]%
        {mckibbin2020global}
\bibfield{author}{\bibinfo{person}{Warwick~J McKibbin} {and}
  \bibinfo{person}{Roshen Fernando}.} \bibinfo{year}{2020}\natexlab{}.
\newblock \showarticletitle{The global macroeconomic impacts of COVID-19: Seven
  scenarios}.
\newblock  (\bibinfo{year}{2020}).
\newblock


\bibitem[\protect\citeauthoryear{Noroozi, Corneanu, Kamińska, Sapiński,
  Escalera, and Anbarjafari}{Noroozi et~al\mbox{.}}{2018}]%
        {article}
\bibfield{author}{\bibinfo{person}{Fatemeh Noroozi}, \bibinfo{person}{Ciprian
  Corneanu}, \bibinfo{person}{Dorota Kamińska}, \bibinfo{person}{Tomasz
  Sapiński}, \bibinfo{person}{Sergio Escalera}, {and}
  \bibinfo{person}{Gholamreza Anbarjafari}.} \bibinfo{year}{2018}\natexlab{}.
\newblock \showarticletitle{Survey on Emotional Body Gesture Recognition}.
\newblock \bibinfo{journal}{\emph{IEEE Transactions on Affective Computing}}
  \bibinfo{volume}{PP} (\bibinfo{date}{01} \bibinfo{year}{2018}).
\newblock
\urldef\tempurl%
\url{https://doi.org/10.1109/TAFFC.2018.2874986}
\showDOI{\tempurl}


\bibitem[\protect\citeauthoryear{Osokin}{Osokin}{2018}]%
        {osokin2018lightweight_openpose}
\bibfield{author}{\bibinfo{person}{Daniil Osokin}.}
  \bibinfo{year}{2018}\natexlab{}.
\newblock \showarticletitle{Real-time 2D Multi-Person Pose Estimation on CPU:
  Lightweight OpenPose}. In \bibinfo{booktitle}{\emph{arXiv preprint
  arXiv:1811.12004}}.
\newblock


\bibitem[\protect\citeauthoryear{Podgorelec, Kokol, Stiglic, and
  Rozman}{Podgorelec et~al\mbox{.}}{2002}]%
        {10.1023/A:1016409317640}
\bibfield{author}{\bibinfo{person}{Vili Podgorelec}, \bibinfo{person}{Peter
  Kokol}, \bibinfo{person}{Bruno Stiglic}, {and} \bibinfo{person}{Ivan
  Rozman}.} \bibinfo{year}{2002}\natexlab{}.
\newblock \showarticletitle{Decision Trees: An Overview and Their Use in
  Medicine}.
\newblock \bibinfo{journal}{\emph{J. Med. Syst.}} \bibinfo{volume}{26},
  \bibinfo{number}{5} (\bibinfo{date}{Oct.} \bibinfo{year}{2002}),
  \bibinfo{pages}{445–463}.
\newblock
\showISSN{0148-5598}
\urldef\tempurl%
\url{https://doi.org/10.1023/A:1016409317640}
\showDOI{\tempurl}


\bibitem[\protect\citeauthoryear{Sammut and Webb}{Sammut and Webb}{2010}]%
        {ref1}
\bibfield{editor}{\bibinfo{person}{Claude Sammut} {and}
  \bibinfo{person}{Geoffrey~I. Webb}} (Eds.). \bibinfo{year}{2010}\natexlab{}.
\newblock \bibinfo{booktitle}{\emph{Logistic Regression}}.
\newblock \bibinfo{publisher}{Springer US}, \bibinfo{address}{Boston, MA},
  \bibinfo{pages}{631--631}.
\newblock
\showISBNx{978-0-387-30164-8}
\urldef\tempurl%
\url{https://doi.org/10.1007/978-0-387-30164-8_493}
\showDOI{\tempurl}


\bibitem[\protect\citeauthoryear{Sun, Wang, Zhao, and Zhang}{Sun
  et~al\mbox{.}}{2020}]%
        {sun2020multi}
\bibfield{author}{\bibinfo{person}{Jun Sun}, \bibinfo{person}{Mantao Wang},
  \bibinfo{person}{Xin Zhao}, {and} \bibinfo{person}{Dejun Zhang}.}
  \bibinfo{year}{2020}\natexlab{}.
\newblock \showarticletitle{Multi-View Pose Generator Based on Deep Learning
  for Monocular 3D Human Pose Estimation}.
\newblock \bibinfo{journal}{\emph{Symmetry}} \bibinfo{volume}{12},
  \bibinfo{number}{7} (\bibinfo{year}{2020}), \bibinfo{pages}{1116}.
\newblock


\bibitem[\protect\citeauthoryear{Toshev and Szegedy}{Toshev and
  Szegedy}{2014}]%
        {A}
\bibfield{author}{\bibinfo{person}{Alexander Toshev} {and}
  \bibinfo{person}{Christian Szegedy}.} \bibinfo{year}{2014}\natexlab{}.
\newblock \showarticletitle{DeepPose: Human Pose Estimation via Deep Neural
  Networks}. In \bibinfo{booktitle}{\emph{Proceedings of the IEEE Conference on
  Computer Vision and Pattern Recognition (CVPR)}}.
\newblock


\bibitem[\protect\citeauthoryear{{Wei}, {Ramakrishna}, {Kanade}, and
  {Sheikh}}{{Wei} et~al\mbox{.}}{2016}]%
        {7780880}
\bibfield{author}{\bibinfo{person}{S. {Wei}}, \bibinfo{person}{V.
  {Ramakrishna}}, \bibinfo{person}{T. {Kanade}}, {and} \bibinfo{person}{Y.
  {Sheikh}}.} \bibinfo{year}{2016}\natexlab{}.
\newblock \showarticletitle{Convolutional Pose Machines}. In
  \bibinfo{booktitle}{\emph{2016 IEEE Conference on Computer Vision and Pattern
  Recognition (CVPR)}}. \bibinfo{pages}{4724--4732}.
\newblock


\bibitem[\protect\citeauthoryear{Yang and Ramanan}{Yang and Ramanan}{2013}]%
        {6380498}
\bibfield{author}{\bibinfo{person}{Yi Yang} {and} \bibinfo{person}{D.
  Ramanan}.} \bibinfo{year}{2013}\natexlab{}.
\newblock \showarticletitle{Articulated Human Detection with Flexible Mixtures
  of Parts}.
\newblock \bibinfo{journal}{\emph{IEEE Transactions on Pattern Analysis \&
  Machine Intelligence}} \bibinfo{volume}{35}, \bibinfo{number}{12}
  (\bibinfo{date}{dec} \bibinfo{year}{2013}), \bibinfo{pages}{2878--2890}.
\newblock
\showISSN{1939-3539}
\urldef\tempurl%
\url{https://doi.org/10.1109/TPAMI.2012.261}
\showDOI{\tempurl}


\bibitem[\protect\citeauthoryear{Zhao, Peng, Tian, Kapadia, and Metaxas}{Zhao
  et~al\mbox{.}}{2019}]%
        {zhaoCVPR19semantic}
\bibfield{author}{\bibinfo{person}{Long Zhao}, \bibinfo{person}{Xi Peng},
  \bibinfo{person}{Yu Tian}, \bibinfo{person}{Mubbasir Kapadia}, {and}
  \bibinfo{person}{Dimitris~N. Metaxas}.} \bibinfo{year}{2019}\natexlab{}.
\newblock \showarticletitle{Semantic Graph Convolutional Networks for 3D Human
  Pose Regression}. In \bibinfo{booktitle}{\emph{IEEE Conference on Computer
  Vision and Pattern Recognition (CVPR)}}. \bibinfo{pages}{3425--3435}.
\newblock


\end{thebibliography}

\end{document}